# Optimizing Over Radial Kernels on Compact Manifolds

Sadeep Jayasumana, Richard Hartley, Mathieu Salzmann, Hongdong Li, and Mehrtash Harandi
Australian National University, Canberra   NICTA, Canberra[*]
sadeep.jayasumana@anu.edu.au

## Abstract

*We tackle the problem of optimizing over all possible positive definite radial kernels on Riemannian manifolds for classification. Kernel methods on Riemannian manifolds have recently become increasingly popular in computer vision. However, the number of known positive definite kernels on manifolds remain very limited. Furthermore, most kernels typically depend on at least one parameter that needs to be tuned for the problem at hand. A poor choice of kernel, or of parameter value, may yield significant performance drop-off. Here, we show that positive definite radial kernels on the unit $n$-sphere, the Grassmann manifold and Kendall's shape manifold can be expressed in a simple form whose parameters can be automatically optimized within a support vector machine framework. We demonstrate the benefits of our kernel learning algorithm on object, face, action and shape recognition.*

## 1. Introduction

Kernel methods in Euclidean spaces have proven immensely successful to address a variety of computer vision problems [18]. The underlying idea is to map input measurements to points in a high-, possibly infinite-, dimensional Hilbert space using a kernel function. Since the mapping is done from a low-dimensional space to a high-dimensional one, it yields a richer representation of the data, and hence typically makes tasks such as classification and clustering easier. However, only positive definite kernels yield a mapping to a valid Hilbert space [18].

Recently, the extension of kernel methods to nonlinear manifolds has drawn significant attention in the computer vision community [8, 10]. This was motivated by the fact that data lying on a Riemannian manifold lack a vector space structure and should therefore not be analyzed with Euclidean methods. Instead of relying on inaccurate tangent space approximations to linearize manifold-valued data, these methods make use of positive definite kernels on the manifolds, which can be thought of as mapping the manifold-valued data to a Hilbert space where Euclidean geometry applies [8, 10]. Therefore, in addition to providing the same benefits as kernel methods in Euclidean spaces, kernel methods on manifolds also let us effectively account for the nonlinear geometry of the data.

Unfortunately, while a wide range of positive definite kernels are known for Euclidean spaces, the variety of such kernels on manifolds remains very limited [10]. It is, however, commonly accepted that the choice of kernel is of great importance for the success of any kernel-based algorithm. In particular, there is a significant amount of literature discussing the influence of the kernel on Support Vector Machines (SVM) [22, 4]. But, nonetheless, there is no established principled way of selecting a kernel for a given task. To the best of our knowledge, the closest solutions to this problem are multiple kernel learning (MKL) [7, 16] and kernel matrix learning [26, 14] approaches. However, the former methods typically need to define a set of useful base kernels, which in itself is an ill-posed problem, and the latter techniques only work in transductive settings.

In practice, for lack of a better choice, many algorithms both in Euclidean spaces [18] and on Riemannian manifolds [10] end up using the Gaussian radial basis function (RBF) kernel. The Gaussian kernel is a radial kernel and, as such, exhibits a number of desirable properties, such as translation invariance, characteristicness and universality [22, 20, 6]. However, it is merely one representative of the wide class of positive definite radial kernels. Furthermore, the Gaussian RBF kernel itself has one parameter, *i.e.*, the bandwidth, that is data-dependent and needs to be tuned. Intuitively, an algorithm that searches over the entire class of radial kernels to find the best-suited kernel for the given problem should perform better than an algorithm that uses a fixed kernel such as the Gaussian RBF kernel.

In this paper, we introduce an algorithm that directly tackles this challenging kernel optimization problem in the context of classification of manifold-valued data. In particular, we consider the case of the unit $n$-sphere, the Grassmann manifold and Kendall's 2D shape manifold, all of which have proven important for diverse computer vision

---

[*]NICTA is funded by the Australian Government as represented by the Department of Broadband, Communications and the Digital Economy and the ARC through the ICT Centre of Excellence program.

This work was supported in part by an ARC grant.

applications. More specifically, we show that radial kernels on these manifolds can be expressed in a simple parametric form. When employed in an SVM framework, the optimal parameters of these kernels can be obtained automatically, thus effectively yielding an algorithm that selects the best kernel among the class of positive definite radial kernels on the manifold. Since this class contains kernels employed in many previous methods, our approach is supposed to perform at least as well as these methods. Our experimental evaluation on object, face, action and shape recognition shows that, in practice, we even significantly outperform these presently known kernels, thus evidencing the importance of learning the best kernel in a principled manner.

## 2. Related Work

There is a huge literature on applications of Euclidean kernel methods to computer vision problems [18]. Among all positive definite kernels, the families of characteristic and universal kernels are of particular interest [22, 20, 6]. Characteristic kernels can be used to analyze probability distributions on a given input space [20]. Universal kernels, on the other hand, can approximate any function on the input space arbitrarily well. Radial kernels are both universal and characteristic [22, 20], as well as translation invariant, and have therefore become very popular. Importantly, radial kernels are the only kind of kernels that can be defined on a generic metric space which only has a distance measure.

The Gaussian RBF kernel in Euclidean spaces is perhaps the most widely used radial kernel in practice. Nonetheless, to obtain good results with this kernel, it is essential to carefully tune the bandwidth of the Gaussian. Although a number of methods have been proposed to automatically determine this parameter [4, 2], most of these methods do not scale well with the number of training samples. Therefore, the most common approach to determining it in practice is to follow an expensive cross-validation procedure with an exhaustive search over a grid. More importantly, the Gaussian RBF kernel is only one specific instance of the class of radial positive definite kernels; other radial kernels could very well be better-suited to address a given problem.

To generalize over the use of a single, fixed kernel, such as the Gaussian RBF, approaches to learning kernel functions, or matrices, have been proposed [16, 14]. The most popular of these approaches is multiple kernel learning (MKL) [7, 16] where the final kernel function is computed as a linear (often conic) combination of base kernels, whose weights are learned from training data. This idea was extended to manifold-valued features in [23]. Despite its success, MKL suffers from the fact that there is no established way to select good base kernels. Moreover, there is no guarantee that the combinations of the chosen kernels cover all possible positive definite (radial) kernels.

In Euclidean spaces, these limitations were addressed by the idea of infinite kernel learning [1], and, more recently, by an approach to learning translation invariant kernels [19]. However, these methods have not gained widespread success, perhaps because they involve sequentially solving complicated optimization problems (a sum of exponentials in [1] and a QCQP in [19]), and thus scale poorly with the number of training samples.

Here, in contrast to the above-mentioned works, we tackle the problem of learning radial kernels on Riemannian manifolds. Kernel methods on manifolds have recently become popular in computer vision. For example, kernels on the Riemannian manifold of symmetric positive definite matrices were introduced in [10]. Kernel methods on Grassmann manifolds were introduced in [8] and [9]. However, all these works use a fixed kernel on the manifold, which may or may not be the best kernel for the problem at hand. Furthermore, the limited number of known positive definite kernels on manifolds [10, 11] increases their risk of being sub-optimal.

In this paper, we show that radial kernels on the unit $n$-sphere, the Grassmann manifold and Kendall's shape manifold can be expressed in a simple linear form. This greatly increases the family of known positive definite kernels on these manifolds. Furthermore, it lets us effectively search for the best kernel for a given problem, with the advantage over [1, 19] of yielding a simpler optimization problem that can exploit available SVM solvers and thus scale well with the number of training samples.

## 3. Background

In this section, we briefly review the three compact manifolds used in this paper. A (topological) *manifold* is a topological space which is locally homeomorphic to the $n$-dimensional Euclidean space $\mathbb{R}^n$, for some $n$ called the *dimensionality* of the manifold.

A *metric space* $(M, d)$ is a set $M$ endowed with a distance function, or a metric, $d(.,.)$. A compact metric space is bounded, meaning that its distance function has an upper bound. A manifold with an appropriate distance function can be treated as a metric space.

### 3.1. The Unit $n$-sphere

The $n$-dimensional sphere with unit radius embedded in $\mathbb{R}^{n+1}$, denoted by $S^n$, is a compact Riemannian manifold. Since it has non-zero curvature, the geodesic distance derived from its Riemannian geometry is a better distance measure on $S^n$ than the usual Euclidean distance. For two points $\mathbf{x}, \mathbf{y} \in S^n$, the geodesic distance $d_g$ is given by

$$d_g(\mathbf{x}, \mathbf{y}) = \arccos(\langle \mathbf{x}, \mathbf{y} \rangle), \quad (1)$$

where $\arccos : [-1, 1] \to [0, \pi]$ is the usual inverse cosine function and $\langle .,. \rangle$ is the Euclidean inner product.

## 3.2. The Grassmann Manifold

The Grassmann manifold $\mathcal{G}_n^r$, where $n > r$, consists of all $r$-dimensional linear subspaces of $\mathbb{R}^n$. It is a compact Riemannian manifold of $r(n-r)$ dimensions. A point on $\mathcal{G}_n^r$ is generally represented by an $n \times r$ matrix $Y$ whose columns store an orthonormal basis of the subspace. The corresponding point on the Grassmann manifold is then given by $\text{span}(Y)$, which we denote by $[Y]$.

While a number of metrics on the Grassmann manifold have been proposed [8], the projection distance is the most popular for computer vision applications [8, 9]. Given two points $[Y_1], [Y_2]$ on $\mathcal{G}_n^r$, represented by matrices $Y_1, Y_2$ having orthonormal columns, it can be expressed as

$$d_P([Y_1], [Y_2]) = 2^{-1/2} \|Y_1 Y_1^T - Y_2 Y_2^T\|_F, \quad (2)$$

where $\|.\|_F$ denotes the Frobenius norm.

## 3.3. The Shape Manifold

Mathematically, the term *shape* refers to the geometric information of an object after removing scale, translation and rotation. In Kendall's formalism [12], a 2D shape is initially represented by the complex $n$-vector containing the 2D coordinates of $n$ landmarks. This vector is then mean-subtracted and normalized to unit length to remove translation and scale. This yields the *pre-shape* $\mathbf{z}$, which lies on the complex unit $n$-sphere $\mathbb{C}S^{n-1}$. The final shape is obtained by identifying all the pre-shapes that correspond to rotations of the same shape. The resulting 2D shape manifold is identified with the complex projective space $\mathbb{C}P^{n-2}$, which is a compact Riemannian manifold.

In the following, $\mathcal{SP}^n$ denotes the $(2n-4)$-dimensional shape manifold generated by $n$ landmarks, and $[\mathbf{z}]$ denotes the shape represented by a pre-shape $\mathbf{z}$. The popular full Procrustes distance between two shapes is given by

$$d_{FP}([\mathbf{z}_1], [\mathbf{z}_2]) = \left(1 - |\langle \mathbf{z}_1, \mathbf{z}_2 \rangle|^2\right)^{\frac{1}{2}}, \quad (3)$$

where $\langle .,. \rangle$ denotes the complex-valued inner product in the complex Euclidean space.

## 4. Radial Kernels on Compact Manifolds

In this section, we characterize the class of positive definite radial kernels on the compact manifolds described above. Although the theory we rely on was introduced by Schoenberg in 1942 [17], it has received very little attention in the computer vision and machine learning communities.

In the following, we use the term *kernel* to indicate a bivariate real-valued function defined on some nonempty set. Let us first give the formal definition of a positive definite (p.d.) kernel [3].

**Definition 4.1.** *Let $\mathcal{X}$ be a nonempty set. A kernel $f : (\mathcal{X} \times \mathcal{X}) \to \mathbb{R}$ is called a **positive definite kernel** if $f$ is symmetric and*

$$\sum_{i,j=1}^n c_i c_j f(x_i, x_j) \geq 0$$

*for all $n \in \mathbb{N}, \{x_1, \ldots, x_n\} \subseteq \mathcal{X}$ and $\{c_1, ..., c_n\} \subseteq \mathbb{R}$.*

In this work, we utilize the following three well-known closure properties of p.d. kernels on a nonempty set [17, 3].

1. If two kernels $k_1, k_2$ are p.d., then so is their conic combination $a_1 k_1 + a_2 k_2$, where $a_1, a_2 \geq 0$.

2. If two kernels $k_1, k_2$ are p.d., then so is $k_1 k_2$, and therefore $k_1^n$, for all $n \in \mathbb{N}$.

3. If all kernels in a pointwise convergent sequence $k_1, k_2, \ldots$ are p.d., then their pointwise limit $k = \lim_{i \to \infty} k_i$ is also p.d.

Let us now define radial kernels on metric spaces.

**Definition 4.2.** *Let $(M, d)$ be a metric space. A kernel of the form $k(x, y) = (\varphi \circ d)(x, y)$, where $\varphi : \mathbb{R}_0^+ \to \mathbb{R}$ is a function, is called a **radial kernel** on $(M, d)$. Furthermore, $k$ is called a **continuous kernel** if $\varphi$ is continuous.*

When multiple distance functions exist in a space $\mathcal{X}$, and the distance function used here is not obvious, we use the terminology *radial kernels with respect to distance $d$* to indicate radial kernels on $(\mathcal{X}, d)$. Our goal is to characterize p.d. radial kernels on a number of compact manifolds. We start with the unit sphere $S^n$.

## 4.1. Radial Kernels on the Unit Sphere

It was shown in [17] that the class of radial kernels that are p.d. on $S^n$ for all finite $n$ is exactly the same as the class of radial kernels that are p.d. on the infinite dimensional Hilbert sphere $S_\mathcal{H}$ (the unit sphere in the Hilbert space $\mathcal{H}$). We now show a complete characterization of this class.

It is well known that the inner product kernel $k_1(.,.) = \langle .,. \rangle$ is p.d. on $\mathcal{H}$ [18, 3]. Therefore, it is p.d. when restricted to $S_\mathcal{H}$; if $\mathbf{x}$ and $\mathbf{y}$ are points on $S_\mathcal{H}$, then $k_1(\mathbf{x}, \mathbf{y}) = \langle \mathbf{x}, \mathbf{y} \rangle$ is p.d. It is also clear that $k_1$ is a radial kernel with respect to the geodesic distance $d_g$ on $S_\mathcal{H}$, since $\langle \mathbf{x}, \mathbf{y} \rangle = \cos(d_g(\mathbf{x}, \mathbf{y}))$.

From the 2$^{\text{nd}}$ closure property of p.d. kernels stated above, it follows that $k_i(\mathbf{x}, \mathbf{y}) = \langle \mathbf{x}, \mathbf{y} \rangle^i$ is also p.d. for all $i \in \mathbb{N}$. It is also trivially p.d. for $i = 0$. Furthermore, using the 3$^{\text{rd}}$ closure property, we may identify two other p.d. kernels $k_{-1} = \lim_{i \to \infty} k_{2i+1}$ and $k_{-2} = \lim_{i \to \infty} k_{2i}$. Therefore, we have now identified the following three types of radial p.d. kernels on $S_\mathcal{H}$:

$$k_i(\mathbf{x}, \mathbf{y}) = \langle \mathbf{x}, \mathbf{y} \rangle^i, \text{ for } i \in \mathbb{N}_0,$$

$$k_{-1}(\mathbf{x}, \mathbf{y}) = \begin{cases} 1 & \text{if } \mathbf{x} = \mathbf{y}, \\ -1 & \text{if } \mathbf{x} = -\mathbf{y}, \\ 0 & \text{otherwise}, \end{cases}$$

and

$$k_{-2}(\mathbf{x}, \mathbf{y}) = \begin{cases} 1 & \text{if } \mathbf{x} = \pm\mathbf{y}, \\ 0 & \text{otherwise}. \end{cases}$$

It is easy to see that $k_{-1}$ and $k_{-2}$ take the forms given above by noting that $|\langle \mathbf{x}, \mathbf{y}\rangle| \leq 1$.

According to the 1st closure property, p.d. kernels on any given set form a convex cone. Berg et al. [3] proved, based on a result by Schoenberg in 1942 [17], that $\{k_i(\mathbf{x}, \mathbf{y})\}_{i=-2}^{\infty}$ spans the cone of all p.d. radial kernels on $(S_\mathcal{H}, d_g)$. We now formally state this result.

**Theorem 4.3.** *A kernel $k : S_\mathcal{H} \times S_\mathcal{H} \to \mathbb{R}$ is radial with respect to the geodesic distance and is p.d. if and only if it admits the form*

$$k(\mathbf{x}, \mathbf{y}) = \sum_{i=-2}^{\infty} a_i k_i(\mathbf{x}, \mathbf{y})$$

*where $\sum_i a_i < \infty$ and $a_i \geq 0$ for all $i$. Furthermore, $k$ is continuous if and only if $a_{-1} = a_{-2} = 0$.*

*Proof.* Schoenberg's original proof for continuous $k$s can be found in [17]. The proof of the complete theorem is given in Theorem 5.3.6 in [3]. □

The theorem stated above completely characterizes the family of p.d. kernels on $S_\mathcal{H}$ that are radial with respect to $d_g$. Since the Euclidean distance on $S_\mathcal{H}$ is a function of only $d_g$, this family is the same as the family of p.d. radial kernels with respect to the Euclidean distance. This is a very strong and useful result, since, as will be shown later, it allows us to derive an algorithm that searches the entire space of p.d. radial kernels to find the best kernel for a given problem on $S_\mathcal{H}$.

Due to the frequent use of normalized image features, many computer vision problems make use of data that lie on unit spheres. The nonlinear geometry of the sphere makes the geodesic distance $d_g$ a better measure than the Euclidean distance. However, many techniques still exploit the Gaussian RBF kernel $\exp(-\gamma\|\mathbf{x} - \mathbf{y}\|^2)$, which relies on the Euclidean distance. In contrast, we make use of Theorem 4.3 to optimize over all possible radial kernels (with respect to both geodesic and Euclidean distances). The resulting kernel can therefore be expected to perform better than the usual Gaussian RBF kernel, since the latter is a member of the family of kernels that we consider. This can be verified by considering the following Taylor series expansion of the Gaussian RBF for $\mathbf{x}, \mathbf{y} \in S_\mathcal{H}$:

$$k_G(\mathbf{x}, \mathbf{y}) = \exp(-\gamma\|\mathbf{x} - \mathbf{y}\|^2) = \exp(-2\gamma(1 - \langle \mathbf{x}, \mathbf{y}\rangle))$$
$$= \exp(-2\gamma) \sum_{i=0}^{\infty} \frac{2^i \gamma^i}{i!} \langle \mathbf{x}, \mathbf{y}\rangle^i.$$

Therefore, the Gaussian RBF kernel can be expressed in the form given in Theorem 4.3.

### 4.2. Radial Kernels on Metric Spaces

We now generalize Schoenberg's result on p.d. kernels on the unit sphere to other compact manifolds with similar geometries by introducing the following theorem.

**Theorem 4.4.** *Let $(M, d)$ be a metric space and $S_\mathcal{H}$ be the unit sphere in a real Hilbert space $\mathcal{H}$. If there exists a function $G : M \to S_\mathcal{H}$ that is a scaled isometry between $(M, d)$ and $(\mathcal{H}, \|.\|)$, then any kernel $k$ of the form*

$$k(x, y) = \sum_{n=-2}^{\infty} a_i k_i(G(x), G(y))$$

*where $\sum_i a_i < \infty$ and $a_i \geq 0$ for all $i$, is p.d. and radial on $(M, d)$. Furthermore, if $G : M \to S_\mathcal{H}$ is surjective, all p.d. radial kernels on $(M, d)$ are of this form.*

*Proof.* If $G$ is a scaled isometry between $M$ and $\mathcal{H}$, a radial kernel on $M$ takes the form $k(x, y) = (\varphi \circ d)(x, y) = \varphi(\lambda\|G(x) - G(y)\|)$ where $\lambda > 0$. Therefore, a kernel $k(.,.) = k_S(G(.), G(.))$ is radial and p.d. on $(M, d)$ if $k_S(.,.)$ is radial and p.d. on $(S_\mathcal{H}, \|.\|)$. Now, $k_S$ is radial and p.d. on $(S_\mathcal{H}, \|.\|)$, or equivalently, on $(S_\mathcal{H}, d_g)$, if it takes the form given in Theorem 4.3. This completes the proof of the first part of the theorem.

If $G$ maps elements of $M$ onto only a subset of $S_\mathcal{H}$, there could be kernels that are p.d. on that subset (and therefore on $M$) but not on $S_\mathcal{H}$. However, this possibility is eliminated if $G : M \to S_\mathcal{H}$ is surjective, which proves the second part of the theorem. □

### 4.3. Radial Kernels on the Grassmann Manifold

We now describe p.d. radial kernels on the Grassmann manifold $\mathcal{G}_n^r$. Given an $n \times r$ matrix $Y$ with orthonormal columns which span the linear subspace $[Y] \in \mathcal{G}_n^r$, we define $\hat{Y} = YY^T/\sqrt{r}$. We use $\langle .,.\rangle_F$ to denote the Frobenius inner product between two matrices.

Note that $Y^TY = I_r$ and $\hat{Y}$ is an $n \times n$ symmetric matrix with Frobenius norm $\|\hat{Y}\|_F = \langle \hat{Y}, \hat{Y}\rangle_F^{1/2} = (\text{tr}(YY^TYY^T)/r)^{1/2} = (\text{tr}(Y^TYY^TY)/r)^{1/2} = 1$. Therefore, $\hat{Y}$ lies on $S^{(n^2+n-2)/2}$, the unit sphere in the real $(n(n+1)/2)$-dimensional Hilbert space of $n \times n$ symmetric matrices endowed with the Frobenius inner product.

Let us now define $G : \mathcal{G}_n^r \to S^{(n^2+n-2)/2} : G([Y]) = \hat{Y}$. It is clear from Eq. 2 that $d_P([Y_1], [Y_2]) =$

$\sqrt{r/2} \|\hat{Y}_1 - \hat{Y}_2\|_F$. Therefore, $G$ is a scaled isometry satisfying the properties specified in Theorem 4.4.

With some mathematical manipulations, it can be seen that $\langle \hat{Y}_1, \hat{Y}_2 \rangle_F$ is always non-negative and hence $\hat{Y}_1 = -\hat{Y}_2$ never occurs. Therefore, $k_{-1}$ coincides with $k_{-2}$ and the series representation in Theorem 4.3 can be further reduced to the form in the following corollary.

**Corollary 4.5.** *A kernel $k : \mathcal{G}_n^r \times \mathcal{G}_n^r \to \mathbb{R}$ is radial with respect to the projection distance and is p.d. if it admits a series representation of the form*

$$k([Y_1], [Y_2]) = \sum_{i=0}^{\infty} a_i \langle \hat{Y}_1, \hat{Y}_2 \rangle_F^i + a_{-1} 1_{\{1\}}(\langle \hat{Y}_1, \hat{Y}_2 \rangle_F),$$

*where $\sum_i a_i < \infty$, $a_i \geq 0$ for all $i$, and $1_A(t)$ is the indicator function defined as*

$$1_A(t) = \begin{cases} 1 & \text{if } t \in A, \\ 0 & \text{otherwise.} \end{cases}$$

Since the dimensionality of $\mathcal{G}_n^r$ is $r(n-r)$, the mapping $G : \mathcal{G}_n^r \to S^{(n^2+n-2)/2}$ above cannot be surjective in general. Thus, the converse of Corollary 4.5 cannot be proved in this manner. Nevertheless, we note that, as was shown by Schoenberg [17] for the case of $S^n$, p.d. radial kernels that do not admit the series expansion above rapidly diminish when the dimensionality of the compact manifold increases. Furthermore, this series representation covers a wide range of p.d. kernels on the Grassmann manifold. In particular, by making use of Taylor series expansions, it can be seen that it covers the projection kernel $k_P([Y_1], [Y_2]) = \langle \hat{Y}_1, \hat{Y}_2 \rangle_F$ [8] and the Gaussian kernel on the projection space $k_{PG}([Y_1], [Y_2]) = \exp(-\gamma \|\hat{Y}_1 - \hat{Y}_2\|_F^2)$, which can be identified as a valid p.d. kernel using a result in [10].

### 4.4. Radial Kernels on the Shape Manifold

We next focus on Kendall's 2D shape manifold. As in Section 3.3, $\mathcal{SP}^n$ denotes the shape manifold generated by $n$ landmarks, $\mathbf{z}$ is a unit norm $n$-complex vector and $[\mathbf{z}]$ is the shape corresponding to $\mathbf{z}$.

In [11], it was shown that the kernel $k_{FP}([\mathbf{z}_1], [\mathbf{z}_2]) = |\langle \mathbf{z}_1, \mathbf{z}_2 \rangle|^2$ is p.d. on $\mathcal{SP}^n \times \mathcal{SP}^n$. Furthermore, it is obviously real-valued. Therefore, there exists a real Hilbert space $\mathcal{H}_1$ and a function $G : \mathcal{SP}^n \to \mathcal{H}_1$ such that $k_{FP}([\mathbf{z}_1], [\mathbf{z}_2]) = |\langle \mathbf{z}_1, \mathbf{z}_2 \rangle|^2 = \langle G([\mathbf{z}_1]), G([\mathbf{z}_2]) \rangle_{\mathcal{H}_1}$ [3]. Note also that $\|G([\mathbf{z}])\|_{\mathcal{H}_1} = \langle G([\mathbf{z}]), G([\mathbf{z}]) \rangle_{\mathcal{H}_1}^{1/2} = |\langle \mathbf{z}, \mathbf{z} \rangle|^{2 \times 1/2} = 1$ for all $[\mathbf{z}] \in \mathcal{SP}^n$. Therefore, $G$ maps 2D shapes to the unit sphere in $\mathcal{H}_1$, denoted by $S_{\mathcal{H}_1}$.

Now, Eq. 3 yields $d_{FP}([\mathbf{z}_1], [\mathbf{z}_2]) = \left(1 - |\langle \mathbf{z}_1, \mathbf{z}_2 \rangle|^2\right)^{1/2} = (1 - \langle G([\mathbf{z}_1]), G([\mathbf{z}_2]) \rangle_{\mathcal{H}_1})^{1/2} = 2^{-1/2} \|G([\mathbf{z}_1]) - G([\mathbf{z}_2])\|_{\mathcal{H}_1}$. Therefore, $G$ is a scaled isometry between $(\mathcal{SP}^n, d_{FP})$ and $(\mathcal{H}_1, \|.\|_{\mathcal{H}_1})$.

As with Grassmann manifolds, since $|\langle \mathbf{z}_1, \mathbf{z}_2 \rangle|^2$ is non-negative, $G([\mathbf{z}_1]) = -G([\mathbf{z}_2])$ never occurs and the series expansion in Theorem 4.4 can be simplified. This lets us write the following corollary to Theorem 4.4.

**Corollary 4.6.** *A kernel $k : \mathcal{SP}^n \times \mathcal{SP}^n \to \mathbb{R}$ is radial with respect to the full Procrustes distance and is p.d. if it admits a series representation of the form*

$$k([\mathbf{z}_1], [\mathbf{z}_2]) = \sum_{i=0}^{\infty} a_i |\langle \mathbf{z}_1, \mathbf{z}_2 \rangle|^{2i} + a_{-1} 1_{\{1\}}(|\langle \mathbf{z}_1, \mathbf{z}_2 \rangle|^2)$$

*where $\sum_i a_i < \infty$ and $a_i \geq 0$ for all $i$.*

In contrast to the case of Grassmann manifolds, the explicit form of the mapping $G : \mathcal{SP}^n \to S_{\mathcal{H}_1}$ above, and even the dimensionality of $\mathcal{H}_1$, are not known. However, $G$ is clearly not surjective. Therefore, there might exist some p.d. radial kernels on $\mathcal{SP}^n$ that do not admit the form given above. However, as before, following Schoenberg's analysis on $S^n$, such kernels diminish rapidly with the dimensionality of $n$. By making use of Taylor series expansions, it can again be shown that this representation covers all p.d. kernels on the shape manifold that have been proposed in the literature. These kernels are the Procrustes Gaussian kernel $k_{FPG}([\mathbf{z}_1], [\mathbf{z}_2]) = \exp(-\gamma(1 - |\langle \mathbf{z}_1, \mathbf{z}_2 \rangle|^2))$ introduced in [11], and the kernel $k_{FP}([\mathbf{z}_1], [\mathbf{z}_2]) = |\langle \mathbf{z}_1, \mathbf{z}_2 \rangle|^2$, which we call the Procrustes kernel.

## 5. Optimizing Over Radial Kernels with MKL

In this section, we show how MKL can be employed to optimize over radial kernels on manifolds.

MKL aims to learn the best kernel function for a given classification task. The most popular approach to MKL consists in expressing the kernel function as a conic combination $\sum_{i=0}^{N} a_i k^{(i)}(x, y)$ of a finite number of base kernels $k^{(0)}, \ldots, k^{(N)}$, and learning weights $a_i \geq 0$ that yield the optimal kernel function for the problem at hand [7, 16]. This is typically done by minimizing an objective function corresponding to the generalization error of an SVM classifier. To avoid over-fitting, Tikhonov regularization, *i.e.*, $\lambda \|\mathbf{a}\|_p$, or Ivanov regularization, *i.e.*, $\|\mathbf{a}\|_p = 1$, are often used.

As was shown in Section 4, radial kernels on the three manifolds we consider can be written as conic combinations of infinitely many monomial kernels of the form $\langle \mathbf{x}, \mathbf{y} \rangle^i$ where $|\langle \mathbf{x}, \mathbf{y} \rangle| < 1$, and the two kernels $k_{-1}$ and $k_{-2}$. As stated in Section 4.1, since $|\langle \mathbf{x}, \mathbf{y} \rangle| < 1$, when $i$ grows, $\langle \mathbf{x}, \mathbf{y} \rangle^i$ approaches $k_{-1}$ for odd $i$s and $k_{-2}$ for even $i$s. From a practical standpoint, this means that, on all three manifolds, high degree monomial kernels do not bring any new information and can thus be ignored. This lets us approximate the infinite series $\sum_{i=-2}^{\infty} a_i k_i$ with the finite sum $\sum_{i=-2}^{N} a_i k_i$. It can easily be verified that this does not affect the radial or p.d. properties of the kernel.

Therefore, we can directly employ any available MKL algorithm to find the best kernel on the manifold by finding the best conic combination of $(N + 1)$ monomial kernels, $k_{-1}$ and $k_{-2}$. In particular, we make use of the SimpleMKL algorithm of [16], which efficiently handles large scale data. In practice, we observed that $N = 10$ is usually enough to obtain good results and therefore employ this value throughout our experiments. Note, however, that as long as it is large enough, the exact value of $N$ is not critical, since the MKL algorithm automatically identifies the kernels that are not useful for classification. This implies that our approach not only is much less restricted than the usual SVMs with a fixed kernel such as the Gaussian RBF kernel, but also does not require any manual tuning of kernel parameters. Even standard MKL methods involve manually defining the base kernels. Our experiments show the benefits of exploiting the rich class of radial kernels in a principled manner.

## 6. Experimental Evaluation

We evaluated our algorithm on a number of problems defined on the three different manifolds considered above. In all our experiments, when Gaussian RBF kernels of the type $k_G(x, y) = \exp(-\gamma f(x, y))$ are used as baselines, we determined $\gamma$ by cross-validation and by performing a grid search over 20 logarithmically spaced values from $10^{-2}$ to $10^2$. Note that our method, on the other hand, does not require any kernel parameter tuning. Where appropriate, we also report conventional MKL results with 10 Gaussian kernels having log-spaced $\gamma$ values from $10^{-2}$ to $10^2$. We refer to this method as MKL with Gaussians.

### 6.1. Classification on the Unit $n$-Sphere

Histogram-based descriptors, such as SIFT, HOG and Bag-of-Words (BoW) features, are widely used in computer vision. Typically, these descriptors are normalized using the $L^1$ or $L^2$ norm, either directly, or block-wise. When $L^2$ normalization is used, the resulting descriptors lie on $S^n$. In the case of $L^1$ normalization, it has been shown that it is often beneficial to apply an element-wise square root to the resulting $L^1$ normalized histogram, which encodes a probability distribution [21]. This representation, once again, yields final descriptors that lie on $S^n$. With block-wise normalization, the descriptors are mapped to a sphere with non-unit radius, which is homeomorphic to $S^n$. Therefore, our method can be used with descriptors that have undergone any of these normalizations.

In the following experiments, we compare our approach to state-of-the-art classification methods on the sphere that use Gaussian RBF or exponential $\chi^2$ kernels. Note that, while commonly used [5, 15], these kernels depend on a parameter that needs careful tuning to obtain good results.

| Method | Accuracy |
|---|---|
| Exp. $\chi^2$ kernel | 82.56 [15] |
| Intersection kernel | 85.07 |
| Gaussian RBF kernel | 81.54 |
| MKL with above 3 kernels | 85.31 |
| MKL with Gaussians | 86.26 |
| Our method | **87.56** |

Table 1: **Object recognition with the Oxford-IIT Pet dataset.** Average recognition accuracies of our method and of other kernel SVM and MKL classification baselines.

#### 6.1.1 Object Recognition

As a first experiment, we utilized the Oxford-IIT Pet dataset [15], which consists of $7,349$ images of cats and dogs of 37 different breeds. Here, we considered the task of classifying cats vs dogs.

To this end, we used protocol of [15], where 100 images of each breed are used for training and the rest for testing. Furthermore, we utilized the same BoW of dense SIFT descriptors with the image layout described in [15] that does not use head annotations or segmentations. We refer the reader to [15] for the details of the descriptor computation. The resulting descriptors are $L^1$ normalized histograms of $20,000$ dimensions, to which we applied an element-wise square root. Therefore, the final descriptors lie on $S^{19,999}$.

We compare our results to those obtained using an SVM with an exponential $\chi^2$ kernel (reported in [15]), a histogram intersection kernel and the usual Gaussian RBF kernel on the same descriptors. We also compare with conventional MKL results obtained with the above three kernels and 10 Gaussian kernels. As can be seen in Table 1, our method outperforms all these baselines.

#### 6.1.2 Hand Sketch Recognition

We next tackled the task of hand sketch recognition, which was shown to be very challenging [5]. To this end, we made use of the Sketch dataset of [5], which comes with pre-computed descriptors. The dataset contains $20,000$ hand sketches of 250 different object classes.

The descriptors provided by [5] are $L^1$ normalized 500-dimensional histograms computed from local line orientation features. As before, we used the square root representation for these histograms and our one-vs-all classification method on $S^{499}$. We evaluated our method using the same protocol as [5], where 3-fold cross-validation was used on 10 subsets of the full dataset.

In Figure 1, we directly compare our results to those reported in [5]. Note that our method outperforms their best method (SVM with the Gaussian RBF kernel) on all test subsets. This once again shows that optimizing over the class of p.d. radial kernels yields better results than approaches that use a fixed type of kernel.

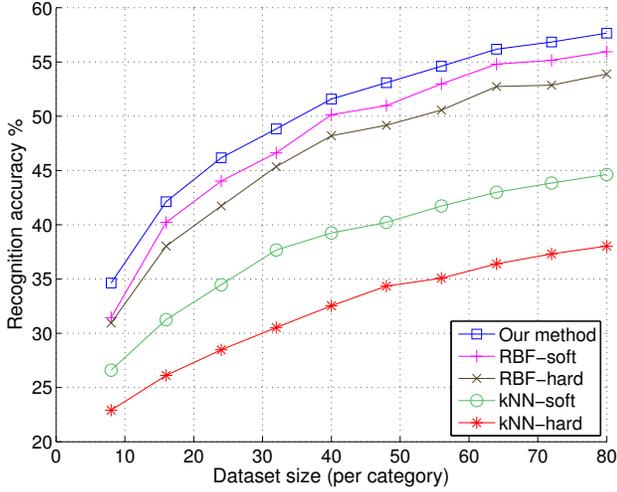

Figure 1: **Hand sketch recognition.** Recognition accuracies for different dataset sizes. The curves for the baselines were reproduced from [5].

## 6.2. Classification on the Grassmann Manifold

We now present our results on $\mathcal{G}_n^r$. Grassmann manifolds are often used to model image sets. More specifically, given a set of descriptors $\{\mathbf{x}_i\}_{i=1}^p$, where $\mathbf{x}_i \in \mathbb{R}^n$ represents an image, an image set can be represented by the linear subspace spanned by the $r(< p, n)$ principal components of $\{\mathbf{x}_i\}_{i=1}^p$. The image set descriptors obtained in this manner lie on the Grassmann manifold $\mathcal{G}_n^r$.

In our experiments, we compare our results to those obtained with state-of-the-art kernel methods on the Grassmann manifold. These methods use the projection kernel $k_P$ [8], and the projection Gaussian kernel $k_{PG}$.

### 6.2.1 Video-based Face Recognition

We first studied the problem of face recognition from video, which can be modeled as linear subspaces [8, 9]. Here, we utilized the challenging YouTube Celebrity dataset [13], which contains video clips of 47 different people.

We used the Viola-Jones face detector to extract face regions from videos and resized them to have a common size of $96 \times 96$. Each face image was then described by a standard Local Binary Patterns (LBP) descriptor and all images corresponding to a video clip by a linear subspace of order 5 extracted from the LBP descriptors. We randomly chose $85\%$ of the total $1471$ image sets for training and the remaining $15\%$ for testing. We report the classification accuracy averaged over 10 such random splits.

We compare our approach with several other kernel methods on $\mathcal{G}_n^r$: Grassmann Discriminant Analysis (GDA) [8], Graph-embedding Grassmann Discriminant Analysis (GGDA) [9], SVM with $k_P$, SVM with $k_{PG}$, and MKL with Gaussians ($k_{PG}$). As can be seen in Table 2, our algorithm outperforms all these kernel methods on the Grassmann manifold.

| Method | YT-Celebrity dataset | Ballet dataset |
|---|---|---|
| GDA [8] | $58.72 \pm 3.0$ | $67.33 \pm 1.1$ |
| GGDA [9] | $61.06 \pm 2.2$ | $73.54 \pm 2.0$ |
| SVM with $k_P$ [8] | $64.76 \pm 2.1$ | $74.66 \pm 1.2$ |
| SVM with $k_{PG}$ [10] | $71.78 \pm 2.4$ | $76.95 \pm 0.9$ |
| MKL with Gaussians | $71.40 \pm 2.0$ | $76.97 \pm 1.2$ |
| **Our method** | $\mathbf{72.00 \pm 1.9}$ | $\mathbf{78.05 \pm 1.0}$ |

Table 2: **Face and action recognition.** Average recognition accuracies of our method compared to other kernel methods on $\mathcal{G}_n^r$.

### 6.2.2 Action Recognition

We next demonstrate the benefits of our algorithm on action classification. To this end, we used the Ballet dataset [25] which contains 44 videos of 8 actions performed by 3 different actors, each video containing different actions performed by the same actor. Action recognition on this dataset is very challenging due to large intra-class variations in speed, clothing and movement.

We grouped every 6 subsequent frames containing the same action, which resulted in 2338 image sets. Each frame was described by a Histogram of Oriented Gradients (HOG) descriptor and 5 principal components were used to represent each image set. Samples were randomly split into two equally sized sets to obtain training and test data. We report the average accuracy over 10 different splits.

We compare our method with the same baselines used in the previous experiment. As evidenced by Table 2, our algorithm yields the best results.

## 6.3. Classification on the Shape Manifold

Finally, we show the benefits of our method on shape classification. Since the state-of-the-art methods attain over $95\%$ accuracy on conventional shape datasets [11], we used two different datasets on which classification using shape only is very challenging: the Oxford-IIT Pet dataset [15] and the Leeds Butterfly dataset [24].

On both datasets, we used the provided segmentation masks to extract 200 landmarks along the contour of each shape with equal arc-length sampling. We compare our method to the SVM and MKL with Procrustes Gaussian kernel $k_{FPG}$ of [11], which was shown to outperform state-of-the-art methods for shape classification. Furthermore, we also report SVM results with the Procrustes kernel $k_{FP}$ and the tangent space Gaussian kernel [11].

### 6.3.1 Butterfly Species Recognition

The Leeds Butterfly dataset [25] contains 832 images from 10 different butterfly classes (species). The task here is to recognize species from shape only. We randomly picked 40 shapes from each class for training and used the rest for testing. In Table 3, we report the average accuracy over 10 such random splits. As before, our learned kernel outperforms fixed-form kernels, including the Procrustes Gaussian kernel which led to state-of-the-art results on shape classification in [11].

| Method | Butterfly dataset | Pet dataset |
|---|---|---|
| SVM with $k_{FP}$ | 57.75 ± 2.0 | 67.48 |
| SVM with $k_{FPG}$ [11] | 60.37 ± 1.6 | 77.34 |
| SVM with Tan. Gauss. kernel [11] | 58.96 ± 1.8 | 75.77 |
| MKL with Gaussians | 60.84 ± 2.0 | 77.24 |
| **Our method** | **63.98 ± 1.6** | **80.87** |

Table 3: **Shape recognition.** Average recognition accuracies of our method compared to other kernel methods on $\mathcal{SP}^n$. Note that the train/test partition on the Pet dataset is fixed and given by [15].

### 6.3.2 Cats vs Dogs Recognition

Finally, we evaluated our shape classification algorithm on the Oxford-IIT Pet dataset [15]. We used pet shape outlines as the sole cue. The results reported in Table 3 show, once again, that learning the best kernel for the task at hand outperforms the use of fixed kernels.

## 7. Conclusion

We have derived characterizations of p.d. radial kernels on the unit $n$-sphere, Grassmann manifold and Kendall's shape manifold. This has allowed us to design an SVM-based algorithm which optimizes over the cone of p.d. radial kernels to find the best kernel that solves a given classification problem on one of these three compact manifolds. By providing a principled way to learn radial kernels on compact manifolds, our algorithm strengthens kernel methods on Riemannian manifolds, which are becoming increasingly popular in computer vision. In the future, we intend to study how this work can be extended to other manifolds.